\pdfoutput=1

\documentclass[11pt]{article}

\usepackage[preprint]{acl}

\usepackage{times}
\usepackage{latexsym}

\usepackage[T1]{fontenc}

\usepackage[utf8]{inputenc}

\usepackage{microtype}

\usepackage{inconsolata}

\usepackage{graphicx}

\usepackage{booktabs}
\usepackage{multirow} 

\title{Does Context Help Mitigate Gender Bias in Neural Machine Translation?}

\author{
\\
\And
\\
\And
Harritxu Gete$^{1,2}$\\
  \And
\\
$^{1}$Vicomtech Foundation, Basque Research and Technology Alliance (BRTA)\\
$^{2}$University of the Basque Country UPV/EHU\\
{\tt \{hgete,tetchegoyhen\}@vicomtech.org}
 \And
Thierry Etchegoyhen$^{1}$\\
\And \\
\And \\
}

\begin{document}
\maketitle
\begin{abstract}
Neural Machine Translation models tend to perpetuate gender bias present in their training data distribution. Context-aware models have been previously suggested as a means to mitigate this type of bias. In this work, we examine this claim by analysing in detail the translation of stereotypical professions in English to German, and translation with non-informative context in Basque to Spanish. Our results show that, although context-aware models can significantly enhance translation accuracy for feminine terms, they can still maintain or even amplify gender bias. These results highlight the need for more fine-grained approaches to bias mitigation in Neural Machine Translation.

\end{abstract}

\section{Introduction}

Neural machine translation (NMT) models tend to exhibit gender bias, originating from their training data \citep{stanovsky-etal-2019-evaluating,saunders-byrne-2020-reducing}. A typical example is the translation of gender-neutral professions in a language like English, into languages with differentiated feminine and masculine forms. In this case, NMT systems often produce translations that reflect gender-stereotypical biases \citep{troles-schmid-2021-extending}. Beyond translation errors, bias perpetuation has a clear negative impact overall.

Several studies have addressed gender bias in NMT, highlighting various sources and manifestations of gender bias in NMT models. For example, \citet{prates2019assessing} and \citet{Rescigno2023} examined gender bias in commercial machine translation systems, revealing a systematic bias towards masculine translation. A variety of approaches have been explored to mitigate these effects, such as data augmentation by swapping gender-specific words \citep{zmigrod-etal-2019-counterfactual,wang-etal-2022-measuring}, or the incorporation of gender tags in the input to guide the translation process \citep{vanmassenhove-etal-2018-getting,corral-saralegi-2022-gender}. The use of context has also been studied as a potential solution, as context-aware models have been shown to significantly enhance translation quality for specific linguistic phenomena, including gender agreement \citep{bawden-etal-2018-evaluating}. 
\citet{basta-etal-2020-towards} and \citet{currey-etal-2022-mt} suggested that context-aware models could help mitigate gender bias in NMT. However, a more detailed study over specific gender categories is still warranted.

In this work, we further explore the role of context in reducing gender bias in NMT, by addressing the following question: does context always help mitigate bias in NMT or can it have bias perpetuation effects? To tackle this question, we studied two specific phenomena related to gender bias. 

First, we evaluated the performance of context-aware models in the translation of stereotypical professions from English into German and French, measuring translation accuracy on gender-based subsets of the data. Our results in this case indicate that, although context-aware models lead to significantly increasing the use of feminine forms, this was achieved mainly for professions that are stereotypically viewed as feminine, thus with limited bias mitigation. 

We then studied the impact on gender bias of non-informative context in Basque to Spanish, i.e., where context lacks gender disambiguating information but nonetheless provides information that may impact the translation. In this case, our results showed significant increases in accuracy for masculine translation options, but notable losses for feminine ones. The use of context was thus detrimental in this case, exacerbating gender biases present in sentence-level models.

Although context can contribute positively to more accurate gender translation, our results highlight the complexity of gender bias in context-aware NMT systems and the need for more fine-grained approaches to mitigate this type of bias.

\section{Experimental Setup}

\subsection{Data}

As training data for our sentence-level baselines, for English-German, we selected the data from the WMT 2017 news translation task; for English-French, we used a mix of publicly available sentence-level parallel data to train baseline models, namely Europarl v7, NewsCommentary v10, CommonCrawl, UN, Giga from WMT 2017 and the IWSLT17 TED Talks \citep{cettolo-etal-2012-wit3}. We then selected the document-level IWSLT17 dataset to train our context-aware models. For evaluation, we selected the contextual subset of MT-GenEval \citep{currey-etal-2022-mt} for both language pairs. 

For Basque to Spanish, we selected the TANDO$^{+}$ \citep{gete2024tando} dataset to train our sentence- and context-level models, and the COH-TGT:GENDER challenge test set for evaluation, which features gender-related context phenomena where the disambiguating information only occurs on the target side. When using models that only have access to the source target context, this test will allow us to measure the impact of non-informative context.

\subsection{Models}

We trained sentence-level baselines and concatenation-based context-aware models. Since the MT-GenEval test set contains only one context sentence in the source language, our analysis is focused solely on models with one source context sentence and no target context (2to1 model). 

Our 2to1 models follow the concatenation approach of \citet{tiedemann-scherrer-2017-neural}, using a Transformer-base architecture \citep{VaswaniSPUJGKP17}, and were trained with MarianNMT \citep{junczys-dowmunt-etal-2018-marian}. 
The embeddings for source, target and output layers were tied and the training was performed using the Adam optimiser \citep{kingma:adam}. Context-aware models were initialised with weights from the sentence-level model.

\subsection{Evaluation}

Since both selected challenge test sets provide contrastive translations, we calculated models accuracy based on their preference for one translation over the other. 
As proposed by \citet{post2024escaping}, we also translated the source sentences and classified the translations as correct or incorrect. We will refer to this type of evaluation as \textit{Generative}. To do this, we first identified correct and incorrect tokens by comparing correct translations with their contrastive counterparts. We then measured accuracy by categorising a translation as successful if it generated a correct token without producing any incorrect ones\footnote{This differs from \citet{currey-etal-2022-mt}, where any case with no incorrect tokens is categorised as successful.}. We also measured incorrect instances, where tokens determined as incorrect are present, and categorised as neutral those cases with neither correct nor incorrect tokens.

\section{Results and Analysis}

\subsection{Stereotypical Professions}

We compare sentence-level models with context-aware models in their ability to translate professions from English into German and French, where, contrary to English, gender-specific forms for professions exist. The test selected for evaluation, MT-GenEval, is balanced in terms of feminine and masculine instances, and contains professions stereotypically viewed as masculine or feminine.

The results for English to German and English to French are shown in Tables~\ref{geneval:total_ende} and~\ref{geneval:total_enfr}, respectively. Overall, context-aware models significantly improved accuracy, in both contrastive and generative evaluations for both language pairs. To assess whether these improvements correlate with a reduction in gender bias, we analysed these results in more details.

\begin{table}[]
\small
\centering
\begin{tabular}{lccc}
\toprule
& \multirow{2}{*}{Contrast.} & Gen.  & Gen.  \\
&  & Correct ($\uparrow$) & Incorrect ($\downarrow$) \\
\midrule
Sent-level & 54.18\%           & 38.00\%          & 36.18\%          \\
2to1  & \textbf{69.27\%}  & \textbf{45.09\%} & \textbf{27.27\%} \\
\bottomrule
\end{tabular}
\caption{Overall accuracy in MT-GenEval (EN-DE)}
\label{geneval:total_ende}
\vspace{10pt}
\begin{tabular}{lccc}
\toprule
& \multirow{2}{*}{Contrast.} & Gen.  & Gen.  \\
&  & Correct ($\uparrow$) & Incorrect ($\downarrow$) \\
\midrule
Sent-level & 53.41\% & 42.95\%          & 39.40\%          \\
2to1  & \textbf{69.24\%} & \textbf{53.41\%} & \textbf{27.93\%} \\
\bottomrule
\end{tabular}
\caption{Overall accuracy in the MT-GenEval (EN-FR)}
\label{geneval:total_enfr}
\end{table}

\begin{table}[t]
\small
\centering
\begin{tabular}{lccc}
\toprule
& \multirow{2}{*}{Contrast.} & Gen.  & Gen.  \\
&  & Correct ($\uparrow$) & Incorrect ($\downarrow$) \\
\midrule
Sent-level & \textbf{92.00\%}           & \textbf{71.09\%}          & \textbf{5.64\%}          \\
2to1  & \textbf{91.45\%}  & 67.27\% & \textbf{6.73\%} \\
\bottomrule
\end{tabular}
\caption{Accuracy over masculine forms (EN-DE)}
\label{geneval:total_ende_masc}
\vspace{10pt}
\begin{tabular}{lccc}
\toprule
& \multirow{2}{*}{Contrast.} & Gen.  & Gen.  \\
&  & Correct ($\uparrow$) & Incorrect ($\downarrow$) \\
\midrule
Sent-level & 92.55\%           & \textbf{76.91\%}          & \textbf{6.00\% }         \\
2to1  & \textbf{96.36\%}  & \textbf{77.09\%} & \textbf{4.73\%} \\
\bottomrule
\end{tabular}
\caption{Accuracy over masculine forms (EN-FR)}
\label{geneval:total_enfr_masc}
\vspace{10pt}
\centering
\begin{tabular}{lccc}
\toprule
& \multirow{2}{*}{Contrast.} & Gen.  & Gen.  \\
&  & Correct ($\uparrow$) & Incorrect ($\downarrow$) \\
\midrule
Sent-level & 16.36\%           & 4.91\%          & 66.73\%          \\
2to1  & \textbf{47.09\%}  & \textbf{22.91\%} & \textbf{47.82\%} \\
\bottomrule
\end{tabular}
\caption{Accuracy over feminine forms (EN-DE)}
\vspace{10pt}
\label{geneval:total_ende_fem}
\begin{tabular}{lccc}
\toprule
& \multirow{2}{*}{Contrast.} & Gen.  & Gen.  \\
&  & Correct ($\uparrow$) & Incorrect ($\downarrow$) \\
\midrule
Sent-level & 14.21\%           & 8.93\%          & 72.86\%          \\
2to1  & \textbf{42.08\%}  & \textbf{29.69\%} & \textbf{51.18\%} \\
\bottomrule
\end{tabular}
\caption{Accuracy over feminine forms (EN-FR)}
\label{geneval:total_enfr_fem}
\end{table}

\begin{figure}
    \centering
    \includegraphics[scale=0.45]{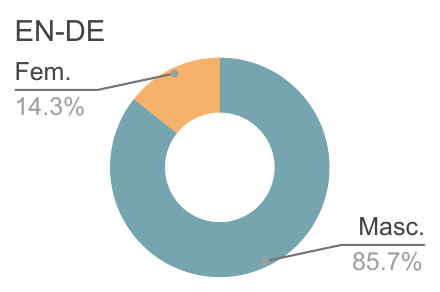}
    \includegraphics[scale=0.45]{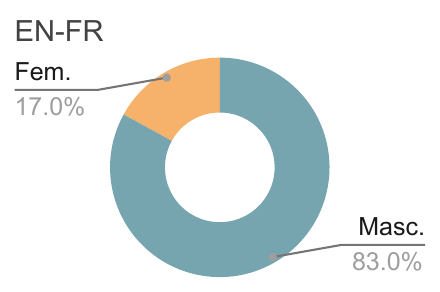}
    \caption{Distribution of sentence-level predictions in stereotipically feminine professions.}
    \label{fig:ster_predictions}
\end{figure}

We manually divided the test into two subsets based on the expected gender and calculated accuracy for each category, with the results shown in Tables~\ref{geneval:total_ende_masc} to \ref{geneval:total_enfr_fem}.
The high scores in the masculine category, and the low scores in the feminine category, in both language pairs with the sentence-level baselines, suggest a significant bias in the data, causing the models to predominantly translate professions into the masculine form. 
Even when translating stereotypically feminine professions, the models generally tend to favour the masculine form as shown in Figure~\ref{fig:ster_predictions}. 

When comparing the results with and without context, improvements are most notable in the feminine category, which increases by about 30 percentage points in both language pairs. In contrast, in the masculine category the increase is less than 4 points for English to French, and there is no significant difference for English to German. It thus seems that context might help mitigate the bias towards masculine forms when translating professions.

We further analysed these results by focusing on professions categorised as stereotypically feminine or masculine, dividing the results into four subcategories based on the type of profession and the expected gender, and assessing accuracy using contrastive evaluation. Results in Tables~\ref{geneval:acc_ende} and~\ref{geneval:acc_enfr} show that, for both language pairs, the largest gains from context originate from the feminine category as expected, but to a much larger degree for professions stereotypically seen as feminine.

The differences in accuracy for the feminine category, between professions classified as feminine and those classified as masculine, thus increased with the use of context. This was the case both in English-German, where the baselines reflected almost no initial differences between the two groups, and English-French where the initial baseline differences were amplified.



These results indicate that context-aware models can help mitigate the bias in favour of masculine translations, significantly increasing the use of feminine forms, but at the same time maintaining or increasing the differences in accuracy between instances that belong to an existing stereotype and those that do not.

\begin{table*}[]
\footnotesize
\centering
\begin{tabular}{lcrllcr}
\cmidrule{1-3} \cmidrule{5-7}
\multirow{2}{*}{Sentence-level} & \multicolumn{2}{c}{Actual gender}                           &  & \multirow{2}{*}{Context-aware} & \multicolumn{2}{c}{Actual gender}                           \\
\cmidrule{2-3} \cmidrule{6-7}
                                & Feminine                    & \multicolumn{1}{c}{Masculine} &  &                                & Feminine                    & \multicolumn{1}{c}{Masculine} \\
      \cmidrule{1-3} \cmidrule{5-7}                          
Stereot. Fem.               & \multicolumn{1}{r}{17.33\%} & 88.67\%                       &  & Stereot. Fem.             & \multicolumn{1}{r}{56.00\% ($\uparrow$ 38.67\%)} & 89.33\% ($\uparrow$ 0.66\%)                       \\
Stereot. Masc.              & \multicolumn{1}{r}{18.00\%}  & 97.33\%                       &  & Stereot. Masc.             & \multicolumn{1}{r}{40.00\% ($\uparrow$ 22.00\%)} & 93.33\%  ($\downarrow$ 4.00\%)   \\
\cmidrule{1-3} \cmidrule{5-7}                 
\end{tabular}
\caption{Accuracy over gender-specific subsets (EN-DE)}
\label{geneval:acc_ende}
\vspace{10pt}
\begin{tabular}{lcrllcl}
\cmidrule{1-3} \cmidrule{5-7}
\multirow{2}{*}{Sentence-level} & \multicolumn{2}{c}{Actual gender}                           &  & \multirow{2}{*}{Context-aware} & \multicolumn{2}{c}{Actual gender}                           \\
\cmidrule{2-3} \cmidrule{6-7}
                                & Feminine                    & \multicolumn{1}{c}{Masculine} &  &                                & Feminine                    & \multicolumn{1}{c}{Masculine} \\
      \cmidrule{1-3} \cmidrule{5-7}                          
Stereot. Fem.               & \multicolumn{1}{r}{21.33\%} & 87.33\%                       &  & Stereot. Fem.              & \multicolumn{1}{r}{49.33\% ($\uparrow$ 28.00\%)} & 96.67\% ($\uparrow$ 9.34\%)                      \\
Stereot. Masc.              & \multicolumn{1}{r}{8.72\%}  & 97.33\%                       &  & Stereot. Masc.             & \multicolumn{1}{r}{30.87\% ($\uparrow$ 22.15\%)} & 97.33\%  ($=$)   \\
\cmidrule{1-3} \cmidrule{5-7}                 
\end{tabular}
\caption{Accuracy over gender-specific subsets (EN-FR)}
\label{geneval:acc_enfr}
\end{table*}

\subsection{Non-informative Context}

In this section, we focus on the effect of introducing context that lacks relevant disambiguating information for the translation into the correct gender. Although this type of analysis is unusual, because standard tests aim to evaluate if a model is able to use contextual information to handle extra-sentential phenomena, context does not always provide relevant information to solve a specific phenomenon, but is still nonetheless present and impacts the actual translation.

For this analysis, we used the COH-TGT:GENDER test of TANDO$^{+}$ for Basque to Spanish translation, where the disambiguating information is on the target side. Specifically, we analysed the parliamentary domain subset of the test, as the results of \citet{gete2024tando} indicate a clear tendency towards masculine translation in this domain. Since we compare a sentence-level model with a 2to1 model that only has access to the source context, we ensure that neither model has access to the disambiguating information on the target side. 

The contrastive evaluation results shown in Table~\ref{tab:tando1} indicate that, as expected, the overall accuracy does not vary with or without context with uninformative context. 
However, when dividing the results by gender category, the accuracy for the masculine category increased to 98\%, while the accuracy for the feminine category decreased from 10\% to 2\%. Using uninformative context actually increased gender bias in this case.

\begin{table}[!htbp]
\small
\centering
\begin{tabular}{lcccc}
\toprule
              && Total & Masculine & Feminine                       \\
\midrule
Sentence-level && 50\%      & 90\%   & 10\%                        \\
2to1           && 50\%      & 98\%   & 2\% \\
\midrule
2to1$^{*}$ - ctxTED    && 50\%      & 88\%   & 12\% \\
\bottomrule
\end{tabular}
\caption{Contrastive accuracy on the parliamentary subsets of COH-TGT:GENDER (EU-ES)}
\label{tab:tando1}
\end{table}

Additionally, we performed a generative evaluation over feminine forms only (Table~\ref{tab:tando2}). In line with the contrastive results, the sentence-level results indicate a clear tendency towards masculine translation in this domain, with 70\% of incorrect instances in the feminine subset. The results also confirm that the use of context exacerbates this tendency, resulting in even fewer correct feminine forms translation. 

\begin{table}[]
\small
\centering
\begin{tabular}{lcc}
\toprule
                    & Correct & Incorrect \\
\midrule
Sentence-level      & 4\%   & 70\%         \\
2to1                & 0\%   & 74\%       \\
\midrule
2to1$^{*}$ - ctxTED   & 4\%   & 70\%         \\
\bottomrule
\end{tabular}
\caption{Generative accuracy on the feminine parliamentary subset of COH-TGT:GENDER (EU-ES)}
\label{tab:tando2}
\end{table}

A possible explanation for these results is that, even if the context does not contain relevant disambiguating information, it may still include domain-related information. In the political domain, which strongly favours masculine translations, the introduction of context might reinforce this bias. 

To test this hypothesis, we evaluated the same sentences from the political domain, but using context from another domain, namely the TED talks subset of the COH-TGT:GENDER test (ctxTED), which also lacks disambiguating information in the source context side.

In the contrastive evaluation, there was a 10 percentage point difference favoring the feminine category over the original 2to1 model, at the expense of the masculine category. In the generative evaluation, the results were identical to sentence-level results. It thus seems that an uninformative context from the same domain can be a factor in actually increasing gender bias.

\section{Conclusions}

This study investigated the impact of context-aware models on mitigating gender bias in NMT, focusing on the translation of professions from English into German and French, as well as translation with uninformative context in Basque to Spanish.

Our results show that, although contextual models can significantly improve the overall translation accuracy on gender-specific terms, this was achieved mainly over stereotypically feminine professions. The use of context actually increased the disparity between stereotypical genders in this case. Non-informative context was also shown to increase gender-related bias in a domain with strong latent bias, when using context from a different domain had no such effects.

These results underscore the need for more comprehensive approaches to bias in NMT, including more specific evaluations over balanced datasets. Novel mitigation techniques and a deeper understanding of the impact of context will also be needed to further address translation bias.

\section{Limitations}
The experiments were conducted exclusively with 2to1 context-aware models. This choice was influenced by the characteristics of one of the test sets, which only provided one context sentence in the source language. As a result, our findings are specific to this type of model, and further research is needed to explore the effects of other context-aware architectures. Expanding the range of test sets and domains would also provide a more comprehensive understanding of biased translation with and without context.

\section{Ethical Considerations}
Our analysis focused solely on binary gender categories, examining translations in terms of masculine and feminine forms. This binary perspective excludes individuals who do not identify with either of these normative genders. Addressing this issue would require developing and incorporating more inclusive linguistic resources and methodologies that recognize and respect non-binary identities.




\bibliography{anthology,custom}

\appendix



\end{document}